\documentclass[conference]{IEEEtran}

\IEEEoverridecommandlockouts

\usepackage{cite}
\usepackage{amsmath,amssymb,amsfonts}
\usepackage{algorithmic}
\usepackage{graphicx}
\usepackage{textcomp}
\usepackage{xcolor}
\usepackage{booktabs}
\usepackage{subcaption}
\def\BibTeX{{\rm B\kern-.05em{\sc i\kern-.025em b}\kern-.08em
    T\kern-.1667em\lower.7ex\hbox{E}\kern-.125emX}}
\begin{document}

\title{Lightweight High-Fidelity Low-Bitrate Talking Face Compression for 3D Video Conference
}


\author{
Jianglong Li$^1$, Jun Xu$^1$, Bingcong Lu$^1$, Zhengxue Cheng$^1$$^\ast$, Hongwei Hu$^2$, Ronghua Wu$^2$, Li Song$^1$$^\ast$\thanks{$^\ast$ indicates Corresponding Authors} \\
$^1$ Shanghai Jiao Tong University, Shanghai, China;  $^2$ AntGroup, Shanghai, China \\
\{jianglongli666, xujunzz, irene\_lu, zxcheng, song\_li\}@sjtu.edu.cn, \{Hongwei.huhw, r.wu\}@antgroup.com
}

\maketitle


\begin{abstract}

The demand for immersive and interactive communication has driven advancements in 3D video conferencing, yet achieving high-fidelity 3D talking face representation at low bitrates remains a challenge. 
Traditional 2D video compression techniques fail to preserve fine-grained geometric and appearance details, while implicit neural rendering methods like NeRF suffer from prohibitive computational costs.
To address these challenges, we propose a lightweight, high-fidelity, low-bitrate 3D talking face compression framework that integrates FLAME-based parametric modeling with 3DGS neural rendering.
Our approach transmits only essential facial metadata in real time, enabling efficient reconstruction with a Gaussian-based head model. 
Additionally, we introduce a compact representation and compression scheme, including Gaussian attribute compression and MLP optimization, to enhance transmission efficiency.
Experimental results demonstrate that our method achieves superior rate-distortion performance, delivering high-quality facial rendering at extremely low bitrates, making it well-suited for real-time 3D video conferencing applications.
\end{abstract}


\begin{figure*}[tbp]
    \centering
    \includegraphics[trim=8mm 3.5mm 7mm 4mm, clip, width=1 \textwidth]{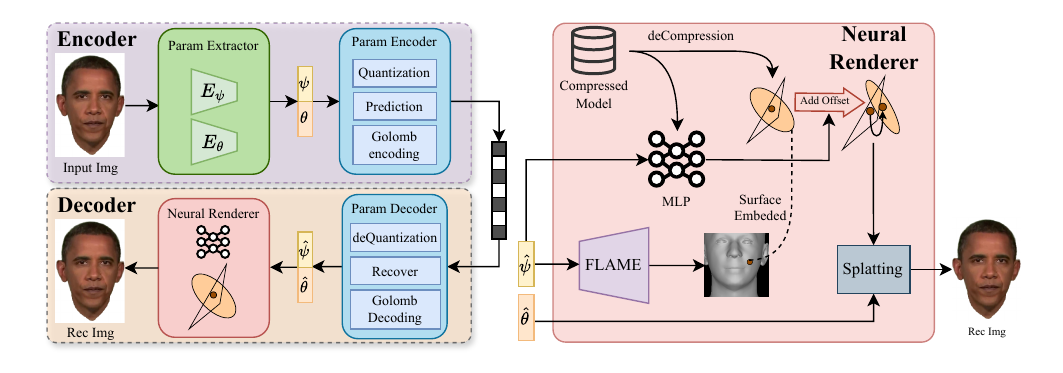}
    \caption{Overall framework. The left panel shows the workflow of the proposed method. The encoder extracts expression parameter $\psi$ and pose parameter $\theta$ from the input image and encodes them into a bitstream, which is then decoded by the decoder to drive the neural renderer and generate the facial image. The right panel illustrates the neural renderer. The compressed facial model is first decoded into Gaussian attributes and MLP. The decoded expression parameters $\hat{\psi}$ are processed by the FLAME model and MLP to obtain Gaussian point positions and attribute offsets. Finally, Gaussian attributes are splatted from the viewpoint provided by the decoded pose parameters $\hat{\theta}$ to reconstruct the facial image.}
    \label{fig:overview}
    \vspace{-1.5em}
\end{figure*}

\section{Introduction}

In recent years, the growing demand for immersive and interactive communication has driven rapid advancements in 3D video conferencing.
However, achieving high-fidelity 3D talking face representation at low bitrates remains a fundamental challenge, as real-time applications must balance visual quality, efficiency, and transmission constraints.

Traditional 2D video compression standards like HEVC \cite{h265} and AV1  \cite{av1} focus on efficient pixel-based encoding for a single viewpoint and do not support the representation of 3D facial geometry. To address this, 3D Morphable Models (3DMM) \cite{3dmm}, such as FLAME \cite{flame}, have been used to encode identity, pose, and expressions, with recent methods \cite{deca,emoca,smirk} enabling real-time parameter extraction. However, these approaches often lack reconstruction fidelity for practical 3D video conferencing. Implicit neural representations, particularly Neural Radiance Fields (NeRF) \cite{nerf} offer high-fidelity rendering but remain too slow and heavy for real-time use, even with compression\cite{gao2022reconstructing,xu2023avatarmav,zhiyuvcip,hu2025},. Recently, 3D Gaussian Splatting (3DGS) \cite{3dgs} has emerged as a more efficient alternative, and several works \cite{flashavatar,GaussianHeadAvatar,gaussianavatar,headgas}  have combined 3DGS with morphable face models to achieve high-quality, real-time talking face rendering.Notably, FlashAvatar \cite{flashavatar} introduces a uniform Gaussian field with spatial offsets to better capture subtle details and non-surface regions, further enhancing visual quality and rendering speed.
Despite these advances, directly applying 3DGS to video conferencing still faces challenges in extracting and updating facial metadata, compressing Gaussian attributes, and balancing compression and fidelity.

Another key challenge is the storage and transmission overhead associated with 3DGS.
Significant redundancy exists among Gaussian attributes, motivating research into compact representations. 
Approaches such as Gaussian pruning \cite{lightgaussian,Lee_2024_CVPR,3dgs_reduce,xie2024mesongs}, codebook-based quantization \cite{lightgaussian,3dgs_reduce,Niedermayr_2024_CVPR,navaneet2023compact3d,xie2024mesongs}, and entropy-based compression \cite{liu2024hemgs} have been proposed to reduce attribute redundancy.
Spatial redundancy can be minimized by structuring Gaussians into 3D \cite{chen2024hac,scaffold} or 2D grids \cite{morgenstern2024compact}, while neural and latent vector representations \cite{eagles2024,chen2024hac,scaffold} further enhance storage efficiency.
There are also some works that focus on training and compressing 3DGS for free-viewpoint video (FVV) \cite{fcgs,4dgc,3dgsstream}, but these methods are not suitable for video conferencing scenarios.

To address these challenges, we propose a high-fidelity, low-bitrate 3D talking face compression framework tailored for 3D video conferencing. Our method integrates FLAME-based parametric modeling with 3DGS to construct an efficient talking face representation. Instead of transmitting raw video frames or dense 3D models, we employ a metadata-driven approach, encoding only essential facial parameters in real time. The decoder reconstructs the talking face using a Gaussian-based head model, significantly reducing bitrate while maintaining high visual quality. Furthermore, we introduce a compact representation and compression scheme for the face model, including Gaussian attribute compression and lightweight MLP-based neural rendering optimization, reducing the face model size from 4.3MB to 0.59MB and further improving transmission efficiency and storage overhead.

In summary, our main contributions are as follows:
\begin{enumerate}
    \item We propose a novel 3D talking face representation for video conferencing, integrating real-time FLAME parameter extraction with 3DGS-based neural rendering for efficient and expressive synthesis.
    \item We introduce a high-efficiency 3D face model compression scheme that achieves more than 7× compression, enabling efficient transmission at session initialization.
    \item Our method outperforms baselines in rate-distortion performance at low bitrates, supports over 170 fps rendering, and shows strong potential for multi-user 3D video conferencing under bandwidth constraints.
\end{enumerate}

\section{Proposed Method}

\subsection{Preliminaries}
\label{subsec:preliminaries}

\textbf{Morphable Head Model.}
FLAME \cite{flame} is a linear statistical model that represents a 3D head mesh using identity parameters $\beta$, pose parameters $\theta$, and expression parameters $\psi$.
The deformed template mesh is computed as:
\begin{equation}
    T_P(\beta, \theta, \psi) = \overline{\mathbf{T}} + B_S(\beta; \mathcal{S}) + B_P(\theta; \mathcal{P}) + B_E(\psi; \mathcal{E}),
\end{equation}
where $\overline{\mathbf{T}}$ is the mean template, and $B_S$, $B_P$, $B_E$ are the shape, pose, and expression blendshapes, respectively.

\textbf{3D Gaussian Splatting.}
3DGS \cite{3dgs} represents a 3D scene as a set of anisotropic Gaussians.
Each Gaussian $G(\boldsymbol{x})$ is defined by its mean $\mu$ and covariance $\Sigma$:
\begin{equation}
    G(\boldsymbol{x})=e^{-\frac{1}{2}(\boldsymbol{x}-\boldsymbol{\mu})^{T} \Sigma^{-1}(\boldsymbol{x}-\boldsymbol{\mu})}.
\end{equation}
The covariance matrix $\Sigma$ can be further decomposed into scaling matrix $S$ and rotation matrix $R$:
\begin{equation}
    \Sigma = RS{S}^T{R}^T.
\end{equation}
The final rendered color $C$ is computed by volume rendering:
\begin{equation}
    C = \sum_{i=1}^{n} c_i \alpha_i' \prod_{j=1}^{i-1} (1 - \alpha_j'),
\end{equation}
where $c_i$ is the color, and $\alpha_i'$ is the blending weight for the $i$-th Gaussian, which is computed from the opacity of the Gaussian.


\subsection{Overall Workflow}
\label{subsec:workflow}
The system consists of a Encoder and Decoder, as illustrated in Fig~\ref{fig:overview}.
Our method encodes a 2D video of a talking face into two components: a pre-trained personalized face model and real-time extracted facial parameters (\S \ref{subsec:face}). Before the meeting, the receiver obtains the face model in advance. We employ a dedicated compression scheme to substantially reduce the model size, thus minimizing the pre-transmission time (\S \ref{subsec:compression}). During the meeting, only the compact facial parameters are transmitted instead of full face images to reduce transmission bitrate.

\textbf{Encoder Side.}
We first utilize the real-time pose and expression estimation network proposed in \cite{smirk} to extract FLAME parameters from the input image:
\begin{equation}
    \psi = E_\psi(I), \quad
    \theta = E_\theta(I), 
\end{equation}
where $E_\psi(I)$ predicts the expression parameters $\psi \in \mathbb{R}^{dim_{\psi}}$, and $E_\theta(I)$ estimates the pose parameters $\theta \in \mathbb{R}^{11}$.
These parameters are then quantized, predicted, and encoded using zero-order Exponential-Golomb coding \cite{gelunbu} to reduce inter-frame redundancy and generate the bitstream.

\textbf{Decoder Side.}
The bitstream is first decoded using zero-order Exponential-Golomb decoding, followed by dequantization to recover the pose and expression parameters. The reconstructed parameters are then fed into the Neural Renderer to synthesize the talking face image.

\subsection{3D Talking Face Representation and Driving}
\label{subsec:face}
Our method leverages the parametric FLAME head model \cite{flame} and the fast rendering capability of 3DGS, making it highly suitable for low-bitrate, real-time video conferencing applications.
Inspired by \cite{flashavatar}, we represent the talking face using a fixed set of Gaussian points and expression-aware Gaussian offsets predicted by an offset network.
Each Gaussian is parameterized as $G = \{\mu, r, s, o, h\}$, where the position $\mu$ is anchored on the FLAME mesh.

The offset network is an MLP $F_\theta$ that takes as input the mesh-based position $\mu_M$ and the expression parameters $\psi$, and outputs the offsets for the Gaussian spatial parameters:
\begin{equation}
    \Delta\mu_\psi,\, \Delta r_\psi,\, \Delta s_\psi = F_\theta(\gamma(\mu_M), \psi),
\end{equation}
where $\gamma$ denotes positional encoding, following \cite{nerf}.
The final Gaussian spatial parameters are then computed as:
\begin{equation}
    \mu_\psi,\, r_\psi,\, s_\psi = (\mu_M \oplus \Delta\mu_\psi,\, r \oplus \Delta r_\psi,\, s \oplus \Delta s_\psi).
\end{equation}
To ensure real-time performance, we train the 3D talking face model using the FLAME parameters (expression $\psi$ and pose $\theta$) extracted by  \cite{smirk}.

In our method, expression parameters account for the majority of the transmitted bitrate, reaching approximately 82\% when using 50-dimensional expression parameters.
Since FLAME expression parameters are derived from principal component analysis (PCA)~\cite{pca}, most facial variations can be effectively captured by only the first few principal components.
Thus, we can reconstruct high-fidelity face images using only a subset of expression parameters, significantly reducing the bitrate with only a minor loss in reconstruction quality.


\subsection{Face Model Compression}
\label{subsec:compression}
\begin{figure}[tbp]
    \centering
    \includegraphics[trim=8mm 4mm 8.3mm 4mm, clip, width=0.475\textwidth]{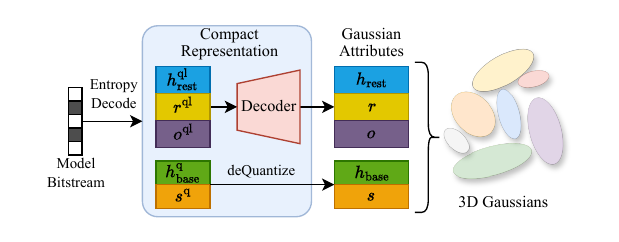}
    \caption{Overview of the compact 3D Gaussian representation.}
    \label{fig:model_compression}
\end{figure}

Although only facial parameters are transmitted during conferencing, the facial model must be sent in advance.
To reduce transmission and storage costs, we compress both components of our face model: the 3D Gaussian attributes and the MLP for attribute offsets.

\textbf{Compact 3D Gaussian Representation.}
Unlike general 3DGS compression methods that prune or restructure Gaussian points, our approach fixes the number of points via UV sampling on the FLAME mesh, making pruning unnecessary.
We therefore focus on compressing the Gaussian attribute values.
The attributes include position $\mu$, base color $h_{\text{base}}$, high-order SH coefficients $h_{\text{rest}}$, scaling $s$, rotation $r$, and opacity $o$.
Since $\mu$ is dynamically generated from the FLAME model and MLP offsets, only $h_{\text{base}}$, $h_{\text{rest}}$, $s$, $r$, and $o$ require compression.

For attributes less sensitive to precision, such as $h_{\text{rest}}$, $r$, and $o$, we follow \cite{eagles2024} and encode them as quantized latent vectors $h_{\text{rest}}^{\text{ql}} \in \mathbb{R}^{\text{dim}_{\text{rest}}}$, $r^{\text{ql}} \in \mathbb{R}^{\text{dim}_{r}}$ and $o^{\text{ql}} \in \mathbb{R}^{\text{dim}_{o}}$ , which are decoded by lightweight neural decoders.
For precision-sensitive attributes, including $h_{\text{base}}$ and $s$, we apply direct quantization to obtain $h_{\text{base}}^q$ and $s^q$, followed by entropy coding.
The final compact model consists of the latent vectors, their decoders, and the quantized $h_{\text{base}}$ and $s$.
All components are further compressed using LZ77 \cite{ziv1978compression,ziv1977universal} for efficient storage and transmission.
The decoding process of the Gaussian attributes is illustrated in Fig~\ref{fig:model_compression}.

\textbf{MLP Compression.}
To minimize the MLP size and avoid overfitting, we employ weight pruning and, if necessary, reduce the number of layers, followed by retraining to recover performance.
The pruned MLP is then quantized to FP16 and compressed with LZ77 for further size reduction.

\section{Experiments}


\begin{figure*}[htbp]
    \centering
    \begin{subfigure}[b]{0.32\textwidth}
        \includegraphics[width=\textwidth]{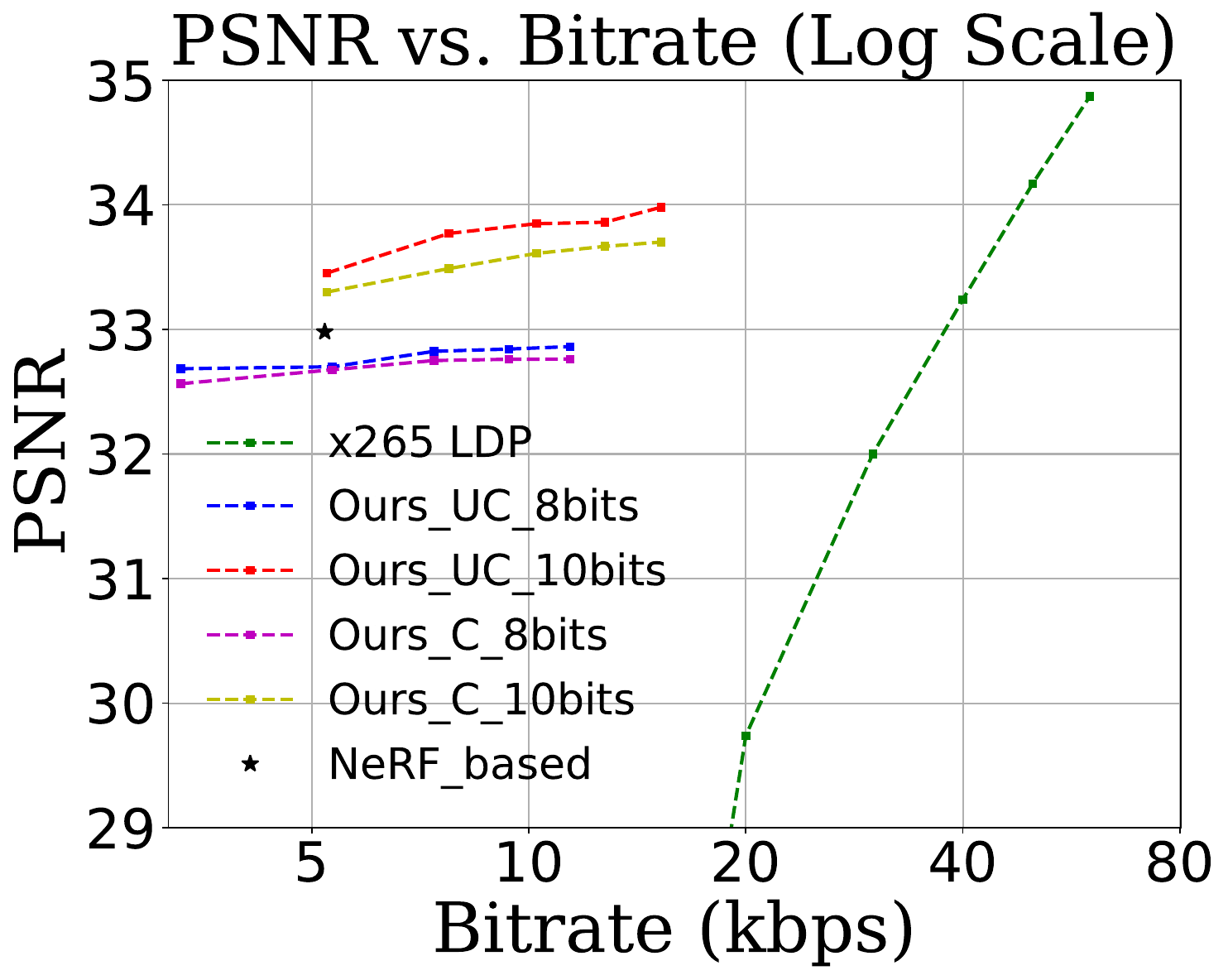}
        \caption{PSNR of x265 LDP at 14kbps is 24.06dB}
    \end{subfigure}
    \begin{subfigure}[b]{0.32\textwidth}
        \includegraphics[width=\textwidth]{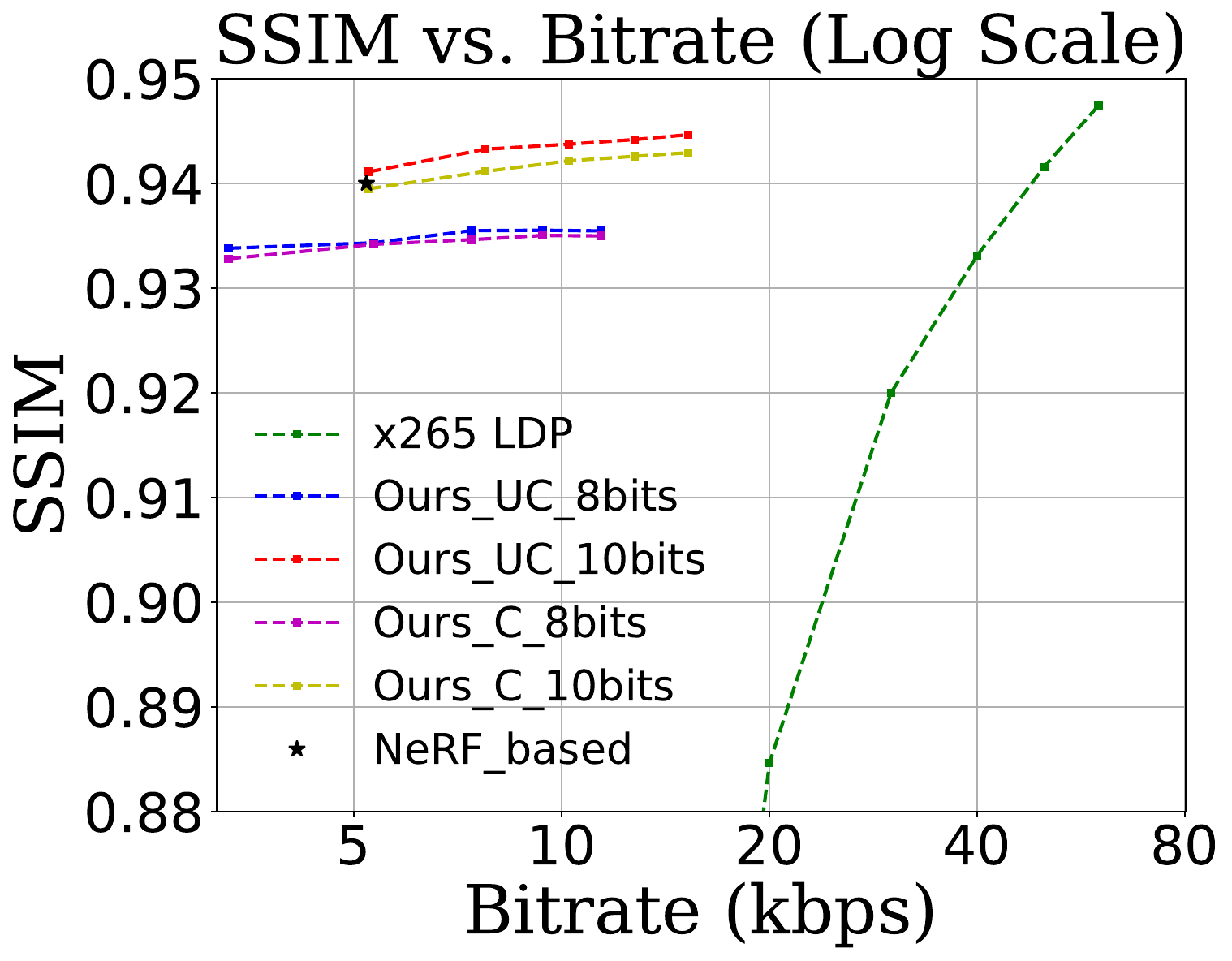}
        \caption{SSIM of x265 LDP at 14kbps is 0.802}
    \end{subfigure}
    \begin{subfigure}[b]{0.31\textwidth}
        \includegraphics[width=\textwidth]{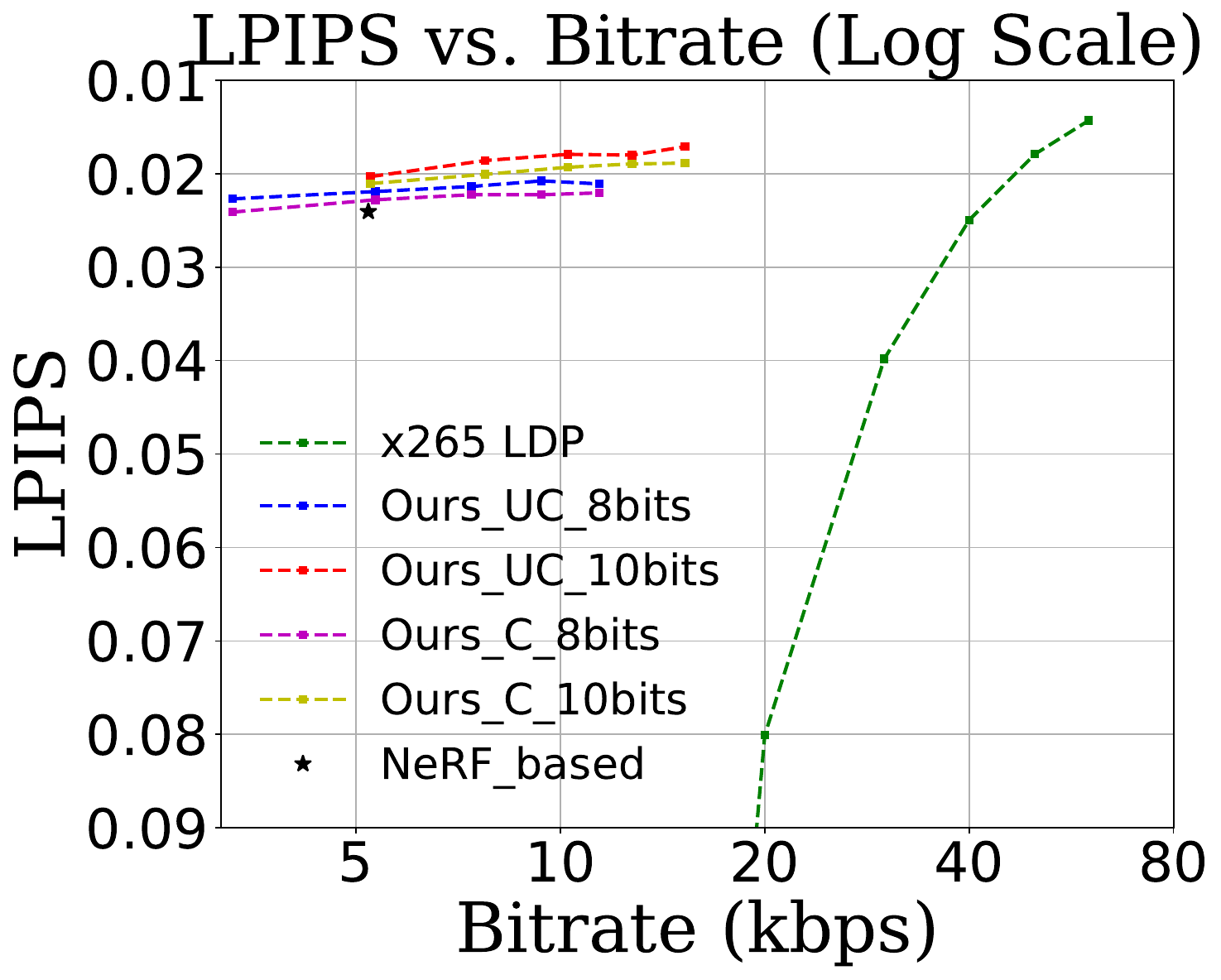}
        \caption{LPIPS of x265 LDP at 14kbps is 0.208}
    \end{subfigure}
    \caption{Performance comparison between our methods, x265 LDP, and NeRF-based method. All videos are encoded at 25 fps. In our methods, ``UC" denotes an uncompressed facial model, while ``C" represents a compressed facial model. ``8bits" and ``10bits" indicate that facial parameters are quantized using 8-bit and 10-bit precision levels, respectively.}
    \label{fig:RD}
    \vspace{-1.5em}
\end{figure*}

\subsection{Implementation Details }
We evaluate our method on datasets provided by \cite{gao2022reconstructing, flashavatar}, featuring videos at a resolution of 512 × 512 with various identities. Each video contains approximately 2,500 frames, with the last 500 frames used for testing and the remaining frames for training. Additionally, we process our own dataset using a methodology similar to that of FlashAvatar \cite{flashavatar}.
By default, we set $ \text{dim}_\psi = [10,20,30,40,50]$, $\text{dim}_{\text{rest}}=1$, $\text{dim}_{r} = 4$, $\text{dim}_o  = 1$.
All the experiments are tested on one single Nvidia GeForce RTX 4090 GPU. 

\subsection{Performance Comparision}

We compare our method with two baselines: the x265 encoder \cite{h265} in LDP mode and a NeRF-based 3D talking face encoding approach \cite{xjibc2024}.
Our method is evaluated in two versions: (1) \textbf{Uncompressed}, where the talking face representation is not compressed, and (2) \textbf{Compressed}, where the representation undergoes compression.
Additionally, we experiment with two quantization precisions for facial parameters (expression $\psi$ and pose $\theta$): \textbf{8-bit} and \textbf{10-bit}. Since our approach is designed for real-time extraction and generation in video conferencing scenarios, whereas methods such as \cite{GaussianHeadAvatar, gaussianavatar, headgas} are intended for reconstruction tasks and cannot accomplish video conferencing tasks, we do not include them in our comparisons.


To ensure fairness, all test videos for each method undergo uniform background replacement.
For the x265 LDP baseline, we use the FFmpeg tool to encode the test videos into H.265 streams at specified bitrates. An example command is: \texttt{ffmpeg -i input -c:v libx265 -x265-params bframes=0 -b:v bitrate -r 25 output.}
We adopt PSNR, SSIM, and LPIPS \cite{lpips} as quantitative metrics to evaluate reconstruction quality.

As shown in Fig~\ref{fig:RD}, our method achieves better rate-distortion performance than x265 LDP at bitrates below 40 kbps. It also outperforms the NeRF-based method under similar bitrates. Our method maintains high rendering fidelity even with reduced quantization precision (10 bit) and lower-dimensional expression parameters (10 dim), as most expressive information lies in the low-dimensional PCA subspace, with diminishing gains from higher dimensions.

\begin{figure}[tbp]
    \centering
    \includegraphics[trim=2mm 2mm 9mm 4mm, clip, width=0.45 \textwidth]{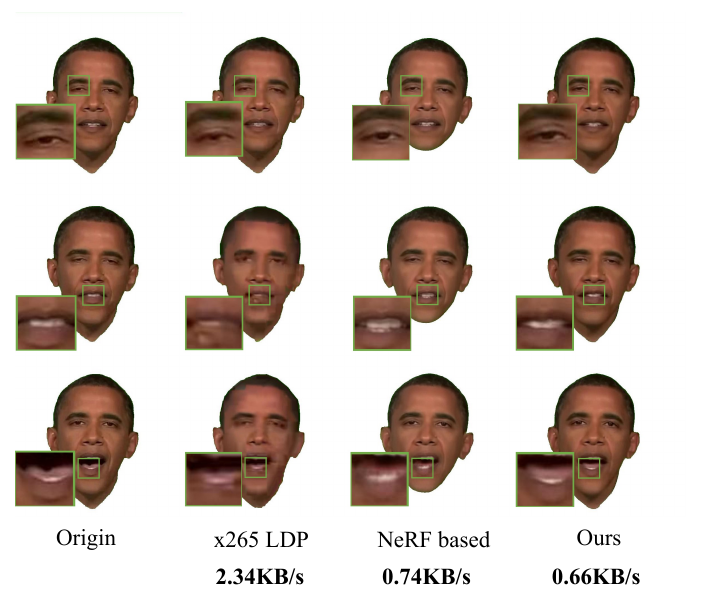}
    \caption{The qualitative comparison results. X265 LDP  exhibits blocking artifacts, NeRF-based method produces unrealistic details in teeth and eyelids regions. Our method generates high-fidelity results under various poses at a lower bitrates.}
    \label{fig:zhuguan}
   
\end{figure}

In Fig~\ref{fig:zhuguan}, the bitrate used by x265 is more than three times higher than that of our method, yet its synthesized results are inferior, exhibiting obvious blocky artifacts.The NeRF-based method operates at a similar bitrate to ours, but fails to faithfully reconstruct fine details such as teeth and eyelids (as indicated by the green boxes), whereas our method preserves these details much better.


\begin{table}[h]
    \caption{Comparison of Model Size and Rendering Speed}
    \label{tab:model_comparison}
    \centering
    \begin{tabular}{lcc}
        \toprule
        Method & Model Size (MB) &  Speed (FPS) \\
        \midrule
        Ours\_compressed   & \textbf{0.59}  & 175 \\
        Ours\_uncompressed & 4.3   & \textbf{181} \\
        NeRF\_based \cite{xjibc2024}              & 180   & 25  \\
        NeRF\_based\_compressed \cite{zhiyuvcip}              & 16   & 25  \\
        \bottomrule
    \end{tabular}
\end{table}

\begin{table}[ht]
    \caption{Ablation Study on Face Model Compression}
    \label{tab:ablation}
    \centering
    \begin{tabular}{lcccc}
        \toprule
        Method & GS attributes   & MLP   & Total Size & PSNR   \\
        \midrule
        Uncompressed        & 2.9 MB  & 1.4 MB  & 4.3 MB & 33.45 dB \\
        + LR          & 0.42 MB & 1.4 MB  & 1.82 MB  & 33.39 dB \\
        + MP      & 0.42 MB & 0.9 MB  & 1.32 MB  & 33.37 dB \\
        + QE        & 0.2 MB  & 0.5 MB  & 0.7 MB  &  33.3 dB \\
        + LZ77            & 0.17 MB & 0.42 MB & 0.59 MB &  33.3 dB \\
        \bottomrule
    \end{tabular}
\end{table}

As shown in Table~\ref{tab:model_comparison}, compared to the NeRF-based approach, our model is significantly smaller and renders much faster.
After compression, our model size is reduced from 4.3 MB to just 0.59 MB, with a rendering frame rate exceeding 170 fps.
In contrast, the compressed NeRF-based model still requires 18 MB and can only render at 25 fps.
A smaller model size leads to shorter pre-transmission times, and a rendering speed exceeding 170 fps allows for real-time rendering for multiple users simultaneously, which is crucial for multi-user 3D video conferencing.

\subsection{Ablation Studies}

We performed an ablation study on the compression components of our 3D facial model, as summarized in Table~\ref{tab:ablation}:

\begin{itemize}
\item \textbf{Latent Representation (LR)}: Uses latent vectors to represent higher-order SH coefficients, rotation, and opacity for Gaussian attributes.
\item \textbf{MLP Pruning (MP)}: Reduces the size of the MLP by pruning its weights.
\item \textbf{Quantization and Entropy Coding (QE)}: Applies quantization and entropy coding to Gaussian attributes and uses FP16 for MLP parameters.
\item \textbf{LZ77 Algorithm}: Further lossless compression.
\end{itemize}

As shown in the rightmost column of Table~\ref{tab:ablation}, each component introduces only minimal PSNR loss.
Overall, our compression reduces the model size from 4.3 MB to 0.59 MB, with just a 0.15 dB drop in rendering quality.

\section{Conclusion}


This paper presents a lightweight, high-fidelity, and low-bitrate 3D talking face compression framework for real-time video conferencing.
By integrating FLAME-based parametric modeling with 3DGS, our approach achieves high-fidelity facial representation at a low bitrate.
Through Gaussian attribute compression and lightweight MLP optimization, we significantly reduce storage and bandwidth requirements without compromising visual quality.
In the future, we plan to build a multi-user 3D video conferencing system and investigate its robustness under challenging network conditions.

\section{Acknowledgment}
This work was partly supported by the NSFC62431015, Science and Technology Commission of Shanghai Municipality No.24511106200, the Fundamental Research Funds for the Central Universities, Shanghai Key Laboratory of Digital Media Processing and Transmission under Grant 22DZ2229005, 111 project BP0719010 and Ant Group Research Fund.
\newpage
\bibliography{reference} 

\end{document}